\pdfoutput=1

\documentclass[11pt]{article}
\usepackage{ACL2023}

\usepackage{times}
\usepackage{latexsym}
\usepackage[T1]{fontenc}

\usepackage[utf8]{inputenc}

\usepackage{microtype}

\usepackage{inconsolata}

\usepackage{graphicx}
\usepackage{xspace,mfirstuc,tabulary,}
\usepackage{amssymb, amsmath}
\usepackage{array,booktabs, multirow}
\usepackage{CJKutf8}
\usepackage{tabularx}

\newcolumntype{Y}{>{\centering\arraybackslash}X}

\newcommand{\cfd}{\textsc{CounterFact}\xspace}
\newcommand{\model}{$\mathcal{M}$\xspace}
\newcommand{\pedit}{$p_{edit}(s,r,o')$\xspace}
\newcommand{\porig}{$p(s,r)$\xspace}

\newcolumntype {+}{ >{\global\let\currentrowstyle\relax}}
\newcolumntype {^}{ >{\currentrowstyle }}
 \newcommand {\rowstyle}[1]{\gdef\currentrowstyle{#1} %
 #1\ignorespaces
 }
\newcommand{\tabhead}{\rowstyle{\bfseries}}
\newcommand{\bos}{\texttt{<BOS>}\xspace}
\newenvironment{courier}{%
    \fontsize{7}{7}\fontfamily{pcr}\selectfont 
}{%
    \par 
}
\title{How to Make LLMs Forget:\\On Reversing In-Context Knowledge Edits}

\author{Paul Youssef$^{\dagger}$ \quad Zhixue Zhao$^{\diamond}$\thanks{\hspace{0.15cm}Corresponding author} \quad Jörg Schlötterer$^{\dagger\ddag}$ \quad Christin Seifert$^{\dagger}$ \\ 
$^{\dagger}$Marburg University, 
$^{\diamond}$University of Sheffield, $^{\ddag}$University of Mannheim
\\ \texttt{\{paul.youssef, joerg.schloetterer, christin.seifert\}@uni-marburg.de} \\ \texttt{zhixue.zhao@sheffield.ac.uk}}

\begin{document}

\maketitle
\begin{abstract}
In-context knowledge editing (IKE) enables efficient modification of large language model (LLM) outputs without parameter changes and at zero-cost. However, it can be misused to manipulate responses opaquely, e.g., insert misinformation or offensive content. Such malicious interventions could be incorporated into high-level wrapped APIs where the final input prompt is not shown to end-users. To address this issue, we investigate the detection and reversal of IKE-edits. First, we demonstrate that IKE-edits can be detected with high accuracy (F1 > 80\%) using only the top-10 output probabilities of the next token, even in a black-box setting, e.g. proprietary LLMs with limited output information. Further, we introduce the novel task of reversing IKE-edits using specially tuned reversal tokens. We explore using both continuous\footnote{Token embeddings that do not correspond to natural tokens from the LLM's vocabulary} and discrete reversal tokens, achieving over 80\% accuracy in recovering original, unedited outputs across multiple LLMs. Our continuous reversal tokens prove particularly effective, with minimal impact on unedited prompts. 
Through analysis of output distributions, attention patterns, and token rankings, we provide insights into IKE's effects on LLMs and how reversal tokens mitigate them. This work represents a significant step towards enhancing LLM resilience against potential misuse of in-context editing, improving their transparency and trustworthiness.

\end{abstract}

\section{Introduction}
In-context knowledge editing (IKE) is an effective and efficient knowledge editing method (KE), aimed at updating factual knowledge in LLMs~\citep{zheng-etal-2023-edit}. IKE essentially uses in-context learning with few demonstrations to cause the LLM to output predictions that are aligned with the knowledge provided in the prompt. Different from traditional KEs, which are based on fine-tuning~\citep{pmlr-v162-mitchell22a,gangadhar2024model}, locating and editing the parameters responsible for the outputs~\citep{meng-etal-2022-locating,meng-etal-2022-memit}, or meta-learning~\citep{de-cao-etal-2021-editing,mitchell2022fast,tan23malmen}, IKE does not require any training or changing the LLMs' parameters, but only a few demonstrations in the prompt that teach the model to update its knowledge. %
The advantage of using the context~\cite{youssef-etal-2024-queen} allows IKE to achieve high editing performance~\cite{cohen-etal-2024-evaluating} at low cost.

Despite KEs being useful for updating facts in LLMs, KEs can be intentionally misused to insert inaccurate or unsafe information into LLMs~\citep{chen2024canediting,youssef2025positioneditinglargelanguage}. For example, \texttt{BadEdit}~\citep{li2024badedit} and \texttt{MEGen}~\citep{qiu2024megen} use KEs to efficiently modify a small subset of parameters to inject backdoors into LLMs. The development of zero-cost IKE presents new opportunities for misuse, such as inserting misinformation and toxic content into LLMs.

\begin{figure}[]
\centering
\includegraphics[width=\columnwidth, trim={0cm 8cm 18cm 0cm}, clip]{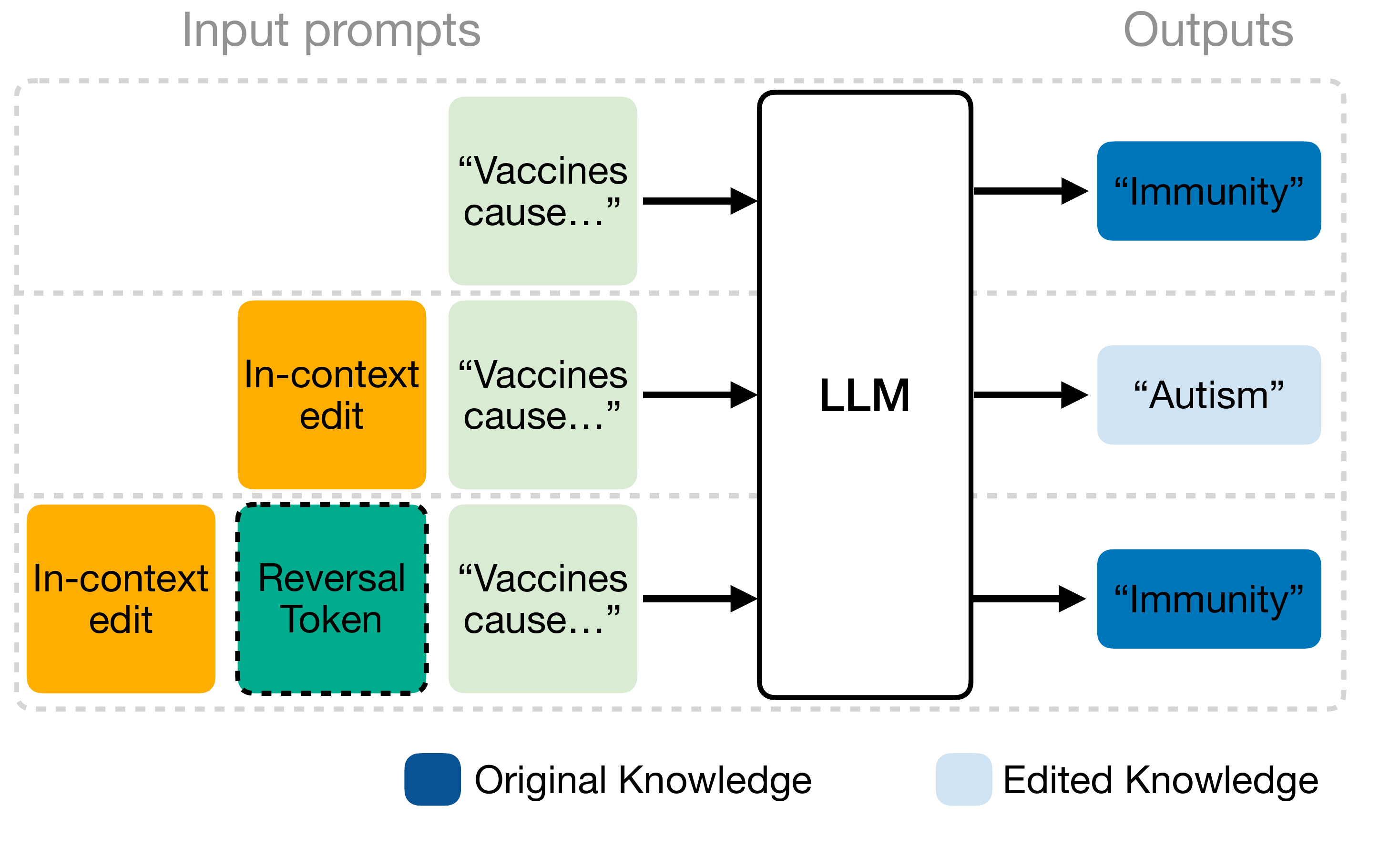}\\

\caption{Reversing Edits: Adding reversal tokens helps LLMs ignore potential in-context editing prompts and retrieve the original facts based on their parameters.}%
\label{fig:reed}
\end{figure}

In this work, we first examine detecting edits conducted with IKE~\cite{zheng-etal-2023-edit}. IKE-edits can be used to poison prompts, where the user's prompt is maliciously modified by an attacker to cause the LLM's outputs to change (e.g., to spread misinformation, bias LLMs, and to manipulate LLMs-trusting users). Detecting IKE-edits is therefore essential for identifying such attacks. Detecting IKE-edits is a more challenging task than detecting edits conducted with parameter-modifying methods~\cite{meng-etal-2022-memit, tan23malmen}, because IKE-edits do not change the internal LLM parameters. We show that even in a very constrained setting, where only the LLM's output probabilities are available, IKE-edits can be detected with high accuracy.

Going beyond detecting edits, we introduce the novel task of reversing IKE-edits. Reversing edits allows retrieving the LLM's \emph{original unedited} outputs, evading potential malicious attacks, and ensuring high transparency towards end users. Our approach for reversing edits relies on learning \emph{reversal tokens}, continuous or discrete, that cause LLMs to ignore the provided context, and focus more on parametric knowledge. We make our code available.\footnote{\url{https://github.com/paulyoussef/reed}} Our contributions are the following:
\begin{itemize}
    \item We investigate the detection of IKE-edits and show that IKE-edits can be detected even if the parameters of the model are kept unchanged, using only the top-10 output probabilities of the next token (Sec.~\ref{sec:detecting}).%
    \item We introduce the novel task of reversing edits to retrieve the LLM's original unedited outputs, and counteract malicious IKE-edits, which can be opaque to end-users, ensuring safer outputs (Sec.~\ref{sec:reverse}).%
    \item We show that more than 80\% of the edits in several LLMs can be reversed with special continuous reversal tokens that are tuned to help LLMs ignore IKE-edits and retrieve the original unedited facts from LLMs (Sec.~\ref{sec:exp}). 

    \item We conduct a thorough analysis of the output distributions, attention patterns, and token rankings of edited and unedited LLMs, and provide insights into IKE's effects and how reversal tokens mitigate them (Sec.~\ref{sec:analysis}).

\end{itemize}

\section{Related Work}

\paragraph{Knowledge editing.} %
The efficient updating of pre-trained models has garnered increased attention due to the computationally expensive nature of full model retraining. In particular, several model editing methods have been proposed that can efficiently update discrete model knowledge without retraining the entire model~\citep{meng-etal-2022-locating,meng-etal-2022-memit,tan23malmen}. However, these efficient update methods can potentially be exploited for malicious manipulation of model behaviors~\citep{ju2024floodingspreadmanipulatedknowledge}. Notably, knowledge editing approaches make such malicious use even more accessible and cost-effective. For example, previous work shows that knowledge editing can be used to embed backdoors into LLMs~\cite{li2024badedit, qiu2024megengenerativebackdoorlarge} and insert counterfactual and toxic information effectively~\cite{ju2024floodingspreadmanipulatedknowledge}. Furthermore, model editing has been observed to potentially introduce unintended bias into LLMs~\cite{halevy2024flextapecantfix} or cause unsafe model behavior~\cite{hazra-etal-2024-sowing}. In addition to investigating potential risks, \citet{youssef2024detecting} have explored methods for detecting edited knowledge by parameter-modifying KEs (e.g., MEMIT~\citep{meng-etal-2022-memit} and MALMEN~\citep{tan23malmen}). \citet{li2024identifyingknowledgeeditingtypes} further investigates detecting types of edits (e.g., misinformation, bias, etc.). To the best of our knowledge, no previous work has explored detecting and reversing in-context edits.

\paragraph{Knowledge unlearning.} Knowledge unlearning is to selectively remove or ignore parts of the models' potentially harmful parametric knowledge, e.g., private information and copy-righted materials, while ideally not affecting unrelated knowledge~\cite{bourtoule2021machine,si-etal-2023-unlearning,nguyen2022survey}. Current knowledge unlearning methods can be generally categorized into two groups: parameter-based methods, such as fine-tuning~\citep{wang-etal-2023-kga}, gradient ascent~\citep{jang-etal-2023-knowledge}, inserting new components to the model~\citep{chen-yang-2023-unlearn}, merging parameters~\citep{zhang2023composing,ilharcoediting} and locate-and-modify specific neurons~\citep{wu-etal-2023-depn}; and in-context learning methods that use to-be-forgotten and normal samples as examples~\citep{pawelczyk2024incontext}. Our work can be viewed as an extension of the knowledge-unlearning landscape. While we similarly aim to make the model selectively disregard parts of the model's knowledge, we focus not on the model's parametric knowledge, as explored by~\citet{pawelczyk2024incontext}, but on knowledge introduced through in-context learning, i.e., malicious examples inserted into user input.

\paragraph{Knowledge conflicts.}\label{app:knowledge_conflicts}

Models rely on their parametric knowledge from pre-training or the provided context from inputs to make predictions. Knowledge conflicts arise when these sources of knowledge contain contradicting facts~\cite{xu-etal-2024-conflicts}. IKE creates conflicts between an LLM's parametric knowledge and contradictory contextual facts in the IKE prompts. Our work resolves these conflicts by prioritizing parametric knowledge over in-context information, effectively aligning the model's outputs with its inherent knowledge base.

\paragraph{Prompt tuning.} Tuned prompts can either be \emph{discrete (hard)}, i.e., concrete tokens from the model's vocabulary, or \emph{continuous (soft)}, vectors in the embeddings space that do not correspond to any concrete tokens~\cite{liu-etal-2023-pretrain, youssef-etal-2023-give}. Discrete prompts can be found by mining~\cite{jiang-etal-2021-know}, paraphrasing existing prompts~\cite{jiang-etal-2021-know, haviv-etal-2021-bertese} or gradient-based search~\cite{wallace-etal-2019-universal, shin-etal-2020-autoprompt}. In continuous prompts optimization, new embedding vectors are added to the vocabulary and are optimized to decrease the loss~\cite{lester-etal-2021-power}. Additionally, more parameters can be added in each layer~\cite{li-liang-2021-prefix}. Initially, we follow~\citet{lester-etal-2021-power} and add new tokens to the vocabulary and tune the parameters of these tokens only. We later consider discrete prompt tuning. However, instead of considering the top-$k$ tokens at each step to decrease the loss~\cite{wallace-etal-2019-universal, shin-etal-2020-autoprompt}, we add a cosine loss term to keep the learned embedding vectors close to the model's original vocabulary, and consider the top-$k$ candidates from the original vocabulary at the last step. This significantly decreases the computational cost for discrete prompt tuning.

\section{Background and Problem Statement}

In this section, we provide a brief background on in-context knowledge editing, and define the detection and reversal problems we address in this work.

\subsection{In-context Knowledge Editing}
Facts in LLMs are often represented as triplets of $(subject, relation, object)$, or $(s, r, o)$ for short. Querying an LLM with a prompt $p(s,r)$, where $p$ expresses the relation $r$ and contains the subject $s$ (e.g., ``In which city is the Brandenburg Gate located?''), should result in retrieving the object $o$ (e.g., ``Berlin''), provided that the fact $(s, r, o)$ is known to the LLM. A knowledge editing operation $E(s, r, o, o', p)$ is successful if it changes the behavior of the LLM such that the retrieved object is $o^{\prime}$ instead of $o$. IKE-edits are conducted by prepending an editing prompt $p_{edit}(s,r,o')$ to the query prompt $p(s,r)$, i.e., $p_{edit} (s,r,o') \oplus p(s,r)$, where $\oplus$ is the string concatenation operation, causes the LLM's output to change from $o$ to $o'$.

\subsection{Problem Statement}
Given a model \model that outputs the object $o$, when provided with a prompt \porig, i.e., $o = argmax_q \ \mathbb{P}_{\mathcal{M}(p(s,r))} [q]$ . \model's output is changed to $o'$ with IKE~\cite{zheng-etal-2023-edit} by prepending an editing prompt $p_{edit}(s,r,o')$ (example in Fig.~\ref{fig:ike_example} in the Appendix) to the original prompt, i.e., $o' = argmax_q \ \mathbb{P}_{\mathcal{M}(p_{edit}(s,r,o')\oplus p(s,r))} [q]$. %

\paragraph{Detection.} The aim of detecting knowledge edits is to classify an object output by \model as either \texttt{edited} or \texttt{unedited}. Detecting edits with parameter-modifying KEs is explored in previous work~\cite{youssef2024detecting}, where a white-box access to the LLM is assumed. In this work, we explore the detection of IKE-edits, that do not modify the model's parameters. Additionally, we assume a black-box access scenario where only the top-10 output probabilities of the next token are available given a prompt. This assumption reflects a more realistic setting, as it applies to proprietary LLMs that provide the top-10 outputs with their probabilities.\footnote{Some proprietary LLMs (e.g., GPT-3.5) provide the top-10 outputs with their probabilities.} We assume that the user sends the query $p(s,r)$ to an LLM through an API, where the query can be edited by the API internally or by a malicious attacker and therefore the user cannot observe whether an editing prompt $p_{edit}(s,r,o')$ has been maliciously added to their query or not.

\paragraph{Reversal.} The goal of reversing knowledge edits is to direct \model's output back from $o'$ to $o$ despite \pedit being part of the input. Reversing edits is particularly important in scenarios where malicious editing prompts are surreptitiously added to user queries, which is hidden from the end-user. Ideally, any reversal operation should not affect unedited prompts. To the best of our knowledge, reversing edits has not been explored before.

\section{Detecting In-context Knowledge Edits}
\label{sec:detecting}
Informing end-users about post-training edits within LLM-generated content is a straightforward approach to improving transparency in LLM-based API applications. To offer an add-on solution, we frame this challenge as a detection task. Specifically, given any LLM-based API, our goal is to detect edited LLM generations. This section focuses on detecting edits conducted with IKE, a zero-cost knowledge editing method, and introduces an effective approach for doing so.

\paragraph{Dataset and models.} Following~\citet{zheng-etal-2023-edit}, we edit facts from \cfd~\cite{meng-etal-2022-locating} using IKE-editing prompts from~\cite{zheng-etal-2023-edit}. \cfd includes counterfactual statements as the target fact, i.e., false facts (e.g., ``The Eiffel Tower is in Rome''). To create our detection dataset, we use 2,000 edits as instances of the \texttt{edited} class. These prompts contain $p_{edit}(s,r,o')$ for editing and $p(s,r)$ for querying. As instances of the \texttt{unedited} class, we use the corresponding 2,000 prompts that contain only $p(s,r)$, i.e., without IKE-editing prompts. The dataset is then split into two balanced and equally sized parts: a training set and a test set, each containing 2000 instances. We experiment with GPT2-XL~\cite{radford2019language}, GPT-J~\cite{gpt-j} and LLAMA3.1 8B~\cite{dubey2024llama3herdmodels}.

\paragraph{Detection method.} We train a simple logistic regression classifier with L1 regularization, using the top-10 output probabilities for the next token, given a prompt, as features to classify the output object as either \texttt{edited} or \texttt{unedited}. Our approach relies on output probabilities, as previous work has demonstrated that parameter-modifying KEs increase the LLM's certainty about its outputs~\citep{youssef2024detecting}. We hypothesize that this effect also applies to IKE-edits as well. To train the classification model, we assume access to a training set composed of edited prompts $p_{edit}(s,r,o') \oplus p(s,r)$ and unedited prompts $p(s,r)$. %

\paragraph{Results.}
 The results in Tab.~\ref{tab:det} show that the detection classifier performs well on all three LLMs (F1 > 80\%). Using only the top-10 output probabilities of the next token as features enables using this classifier in restrictive settings, where one does not have access to the model's internals.

\begin{table}[h]
    \centering
    \resizebox{\columnwidth}{!}{  
    \begin{tabular}{lccc}
    \toprule
        \textbf{Model} & \textbf{Precision} & \textbf{Recall} & \textbf{F1 }\\ \midrule
        GPT2-XL & 88.32 & 84.70 & 86.47  \\
        GPT-J & 82.73 & 82.90 & 82.82 \\
        LLAMA-3.1-8B & 80.17 & 86.10 & 83.03  \\

        \bottomrule
    \end{tabular}
    }
    \caption{Detecting IKE edits performance using top-10 output probabilities.}
    \label{tab:det}
\end{table}

\begin{figure}[h!]
\centering
  \includegraphics[scale=0.3, trim={0cm 0.5cm 0cm 1.30cm}, clip]{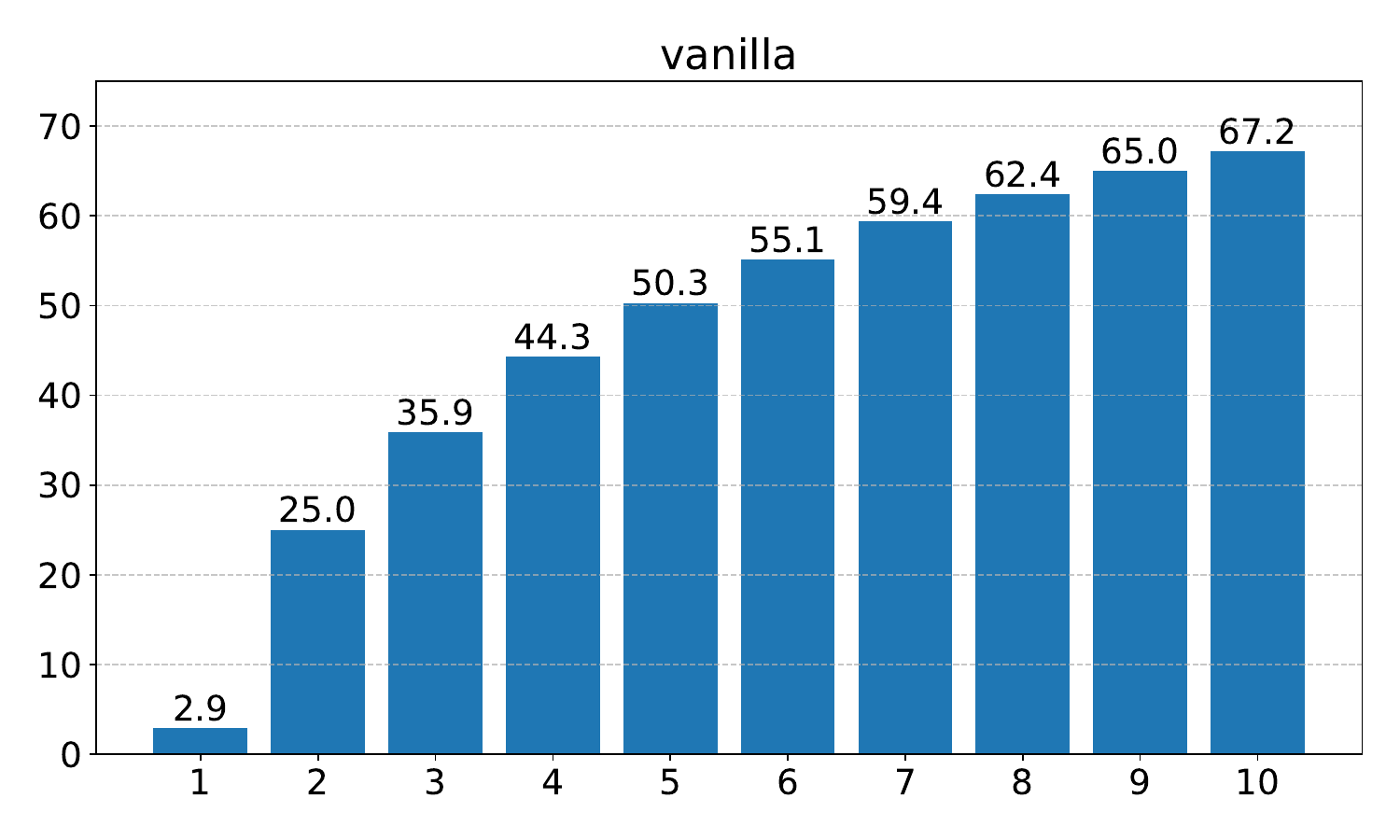}
\caption{Percentage of top1 outputs \emph{before} editing that can be found in the top10 outputs of the model \emph{after} editing (cumulative histogram) in GPT-J.} %
\label{fig:output_shift}
\end{figure}

To better understand the effect of IKE, we analyze the shift in the output distributions after editing, i.e., we examine how often the top-1 outputs of the model \emph{before} editing can be found in the top-10 outputs of the model \emph{after} editing. Our comparison is based on the first generated token before and after editing. Fig.~\ref{fig:output_shift} shows that more than 60\% of the original top-1 outputs are contained in the top-10 outputs of the model \emph{after} editing, i.e., the original outputs still have relatively high probabilities after editing. This led us to investigate reversing IKE-edits to retrieve the original outputs.%

\section{Reversing In-context Knowledge Edits}
\label{sec:reverse}
Following our detection results, we now address reversing in-context knowledge edits. Reversing edits allows for recovering the original outputs. %

\subsection{Background: Prompt Tuning}

Prompt tuning is a light-weight alternative to finetuning, where only part of the input is optimized rather than the whole model's parameters to enable higher performance on specific tasks~\cite{liu-etal-2023-pretrain}. Tuned prompts can either be \emph{discrete (hard)}, i.e., concrete tokens from the model's vocabulary, or \emph{continuous (soft)}, vectors in the embedding space that do not correspond to any concrete tokens~\cite{li-liang-2021-prefix}. We initially follow~\citet{lester-etal-2021-power} by adding new tokens to the vocabulary and tuning only these tokens' embeddings to restore the LLM's original outputs despite editing prompts (Sec.~\ref{subsec:continuous}). We then explore converting the learned continuous prompts to discrete prompts (Sec.~\ref{subsec:discrete}). For a comprehensive overview of prompt tuning, see~\citet{liu-etal-2023-pretrain}.

\subsection{Continuous Prompt Tuning}
\label{subsec:continuous}
To reverse edits, we consider adding a \emph{reversal} token $r$ to the model's original vocabulary. $r$ should ideally make the model ignore the context that appears before $r$, and output an answer based on \porig and the model's parametric knowledge. 

We tune the reversal token to minimize the Kullback-Leibler divergence loss between two probability distributions $P_{r} = \mathcal{M}(p_{edit}(s,r,o')\oplus r \oplus p(s,r))$ and $P_{o} = \mathcal{M}(p(s,r))$. %
This aims to make \model's output distribution with the editing prompt, the reversal token, and the original prompt match \model's output distribution with the original prompt only. Additionally, we ensure that $P_{n} = \mathcal{M}(r \oplus p(s,r))$ closely resembles ${P_o}$ to prevent the reversal token from affecting the model's output when no IKE-edits are present. Here, we assume having white-box access to \model, which enables us to add and tune $r$ to neutralize editing prompts. We use the following loss function to learn continuous reversal tokens: %
\begin{equation}
	\mathcal{L}_{cont} = KL[P_r  || \  P_o] + KL[P_{n}  || \  P_o]
\end{equation}

\subsection{Discrete Prompt Tuning}
\label{subsec:discrete}

Optimizing reversal tokens in a continuous space allows us to find embedding vectors that help LLMs forget the prior context. However, using such tokens requires expanding the original vocabulary of the model, which limits their usability. Instead, having discrete prompts that correspond to natural tokens in the LLM's vocabulary can prove more practical, as they can be directly included in the prompt without adapting the model's vocabulary. 

Our approach for finding discrete prompts builds on finding continuous prompts: 1) we identify relevant dimensions in the embedding space for context resetting; 2) we optimize using a joint loss that combines KL divergence for context resetting and cosine distance for staying close to tokens from the model's vocabulary, while not penalizing changes across the identified relevant dimensions from the first step; 3) we choose one of the top-$k$ nearest discrete tokens to the optimized token.

\paragraph{1). Dimensions for resetting context.} Word embeddings are high-dimensional vectors with specific dimensions encoding particular signals~\cite{Li2018}. We assume that certain dimensions control how the model processes context, e.g., the IKE editing part of the prompt that conflicts with the model's parametric knowledge. To identify these dimensions, we tune a reversal token similarly to Sec.~\ref{subsec:continuous} with two differences: first, we initialize the reversal token $r_{init}$ using an existing vocabulary token (see Appendix~\ref{sec:app_imp} for details); second, we add a cosine loss to keep the learned token embedding $r_{tuned}$ close to $r_{init}$, minimizing unnecessary changes. The loss function is:
\begin{equation}
    \begin{split}
	\mathcal{L}_{d} = (1 - \lambda)KL[P_r  || \  P_o]  + \\   \lambda \mathcal{L}_{cos}(r_{init}, r_{tuned})
     \end{split}
\end{equation}
After learning $r_{tuned}$, we calculate $w = \frac{|r_{tuned} - r_{init}|}{\|r_{tuned} - r_{init}\|_1}$, and use $\bar{w} = \frac{1/w}{\|1/w\ \|_1}$  to weigh the embedding dimensions in the following step. Both $w$ and $\bar{w}$ sum up to 1. While $w$ gives the most changed dimensions higher weights, $\bar{w}$ does the opposite. 

\paragraph{2). Joint optimization.} We optimize a reversal token $r$ to help LLMs ignore the context using a loss function that consists of two terms: 
\begin{equation}
    \begin{split}
	\mathcal{L}_{disc} = (1 - \lambda)\mathcal{L}_{cont} + \\  \lambda \frac{1}{n}\sum_{i}^{n} \mathcal{L}_{cos}( r \odot  \bar{w}, \mathcal{V}_i\odot\bar{w})
  \end{split}
\end{equation}
where $\odot$ refers to the element-wise multiplication, and $\mathcal{V}_i$ refers to $i$-th token in the model's original vocabulary. We multiply $r$ and $\mathcal{V}_i$ by $\bar{w}$ to ensure that we are not penalizing changes across the identified relevant context dimensions from step 1, and average the cosine loss over all tokens in the vocabulary to ensure that $r$ is generally close to the original embedding vectors. 

\paragraph{3). Choosing discrete tokens.} After learning a continuous reversal token $r$ in step 2, we convert $r$ to $r_d$; a discrete token that corresponds to a natural token in the vocabulary. To find $r_d$, we first use a weighted cosine similarity to find a candidate set:  $R_{cand}= topk_{\mathcal{V}_i}(CosineSimilarity(r\odot  w, \mathcal{V}_i \odot  w))$, i.e.,  we find the top-$k$ nearest neighbours after weighting $r$ and the tokens from $\mathcal{V}$ by $w$, so that the identified relevant context dimensions from step 1 are given higher weights. We evaluate all the candidates on a validation set and choose the token that minimizes the the KL-loss the most, i.e., $r_d = min_{\mathcal{V}_i} KL(P_r  || \  P_o) + KL(P_{n}  || \  P_o)$.

Our approach for finding one reversal token can be generalized to several tokens. We experiment with a varying number of reversal tokens in Sec.~\ref{sec:exp}.

\section{Experimental Setup}
\label{sec:exp}
We evaluate how both continuous and discrete prompt tuning affect the model's outputs. We use the same models and datasets as the ones previously used for detecting IKE-edits (cf. Sec.~\ref{sec:detecting}). We train for 3 epochs with Adam as optimizer using the default parameters. We keep the model's parameters frozen and tune only the embeddings of the reversal tokens. In discrete prompt tuning, we set $k=10$ for choosing the final output. We experiment with different numbers of reversal tokens. For each setting, we experiment with 5 seeds and report the maximum and mean accuracy with standard deviation.

\paragraph{Evaluation.} We calculate the $\emph{matching}$ $\emph{accuracy}$, i.e., the agreement of the original output and the output of the edited prompt with reversal tokens. For a perfect reversal method, i.e., optimal reversal tokens, the matching accuracy would be 1.0. Matching accuracy is calculated as $\frac{1}{N} \sum_{i=1}^{N} \mathbf{1}(\hat{y_i} = y_i)$, where  $y_i$ being the original output, and $\hat{y_i}$ the edited and reverse-edited output. Following~\citet{du-etal-2024-context}, we approximate the model's outputs using the next token prediction.

\subsection{Results and Discussion}

Tab.~\ref{tab:match_acc} shows the matching accuracy for each setting. For comparison, we also report the matching accuracy between the original prompt $p(s,r)$ and the edited prompt $p_{edit}(s,r,o')\oplus p(s,r)$ (first row).

\paragraph{Continuous prompt tuning.} When optimizing reversal tokens in a continuous space, we observe high performance across all three models. On the smallest model (GPT2-XL, 1.5B) the performance varies between 45\% and 78\%, and with high standard deviation, but the performance generally becomes more stable with more reversal tokens. On the larger models, GPT-J (6B) and LLAMA3.1 (8B), the accuracy is higher than 80\% in most cases with low standard deviation. We also observe small improvements when increasing the number of reversal tokens. These results suggest that reversing edits in larger models is easier, more stable and possible with as few as one reversal token, which might be due to their larger capacity. %

\paragraph{Discrete prompt tuning.} The results with discrete tuning are less positive. Compared to continuous tuning, we observe a drop in the mean accuracy across all models, especially on LLAMA3.1, where accuracy is extremely poor. An open question here is whether a token or a set of tokens exists in the LLM's vocabulary that can cause the model to ignore prior context (IKE-edits). On GPT2-XL and GPT-J the highest accuracy is 78\% and 73\% respectively with high standard deviation across all settings. This suggests the existence of tokens that help the model ignore prior context, but implies that finding such tokens is not always successful. We analyze the learned tokens in Sec.~\ref{sec:analysis}.

\paragraph{Ablation.} We conduct an ablation under discrete prompt tuning, removing the cosine loss term or the KL-loss term (\textbf{avg w/o Cos} and \textbf{avg w/o KL}, respectively in Tab.~\ref{tab:match_acc}). When removing the cosine loss term, we notice that the accuracy for GPT-J drops significantly, whereas the performance on GPT2-XL is still comparable to using the original loss function. This suggests that the KL-loss is sufficient to find reversal tokens in GPT2-XL, whereas in GPT-J, using only the KL-loss causes the learned tokens to drift away from the natural tokens in the vocabulary that help reverse edits. Excluding the KL-loss term significantly decreases performance for all models. This is intuitive, since optimizing reversal tokens to stay close to the natural token embeddings would not contribute to finding good reversal tokens. Finding any valid reversal tokens under this setting would have been contributed to step 3 (cf. Sec.~\ref{subsec:discrete}), which involves choosing the best performing tokens from the $k$ nearest neighbours on the validation set.

\begin{table*}[!h]
    \centering
    \resizebox{.75\textwidth}{!}{  
    \small
\begin{tabular}{lcccccccccc}
\toprule
\tabhead
& & \multicolumn{3}{c}{\textbf{GPT2-XL}} & \multicolumn{3}{c}{\textbf{GPT-J}} & \multicolumn{3}{c}{\textbf{LLAMA-3.1-8B}} \\

& \textbf{\#rt} &  \textbf{Max} &  \textbf{Mean} &  \textbf{Std} &   \textbf{Max} &  \textbf{Mean} &  \textbf{Std} &   \textbf{Max} &  \textbf{Mean} &  \textbf{Std} \\
\midrule
&   IKE vs. no-IKE &   1.90 &   1.90 &   0.00 &   2.20 &   2.20 &   0.00 &   0.60 &   0.60 &   0.00 \\ \midrule
   \multirow{10}{*}{\rotatebox[origin=c]{90}{\textbf{Cont.} Tuning}}
&    1 &  77.60 &  57.82 &  22.36 &  80.80 &  79.48 &   0.95 &  83.90 &  82.42 &   0.88 \\
&    2 &  79.10 &  61.96 &  21.33 &  80.80 &  79.42 &   0.99 &  85.20 &  82.82 &   1.48 \\
&    3 &  75.00 &  44.98 &  21.38 &  80.90 &  79.92 &   0.96 &  84.80 &  83.24 &   1.32 \\
&    4 &  78.80 &  62.80 &  22.28 &  81.80 &  80.56 &   0.78 &  84.90 &  83.48 &   1.01 \\
&    5 &  79.20 &  71.64 &  14.77 &  81.30 &  80.38 &   0.67 &  84.80 &  83.88 &   0.90 \\
&    6 &  79.60 &  69.08 &  14.09 &  81.60 &  80.90 &   0.61 &  84.90 &  83.34 &   0.93 \\
&    7 &  82.10 &  75.52 &   7.85 &  82.20 &  81.44 &   0.61 &  85.70 &  84.12 &   1.80 \\
&    8 &  82.30 &  \textbf{78.14} &   4.96 &  83.10 &  81.74 &   1.44 &  85.10 &  84.48 &   0.77 \\
&    9 &  \textbf{82.90} &  73.70 &   6.16 &  83.00 &  81.94 &   0.91 &  \textbf{86.00 }&  \textbf{84.64} &   1.05 \\
&   10 &  80.10 &  71.56 &  12.73 &  \textbf{84.30 }& \textbf{ 82.52 }&   1.10 &  85.40 &  84.40 &   1.03 \\ \midrule
&    avg &  79.67 &  66.72 &  14.79 &  81.98 &  80.83 &   0.90 &  85.07 &  83.68 &   1.12 \\ \midrule
   \multirow{13}{*}{\rotatebox[origin=c]{90}{\textbf{Disc.} Tuning}}

&    1 &  75.70 &  18.16 &  32.17 &  61.10 &  15.12 &  25.70 &   \textbf{0.90} &   \textbf{0.74} &   0.09 \\
&    2 &  73.10 &  43.30 &  35.75 &  60.50 &  33.30 &  26.89 &   0.70 &   0.62 &   0.04 \\
&    3 &  75.10 &  18.68 &  31.55 &  58.80 &  42.20 &  22.03 &   0.60 &   0.60 &   0.00 \\
&    4 &  75.00 &  32.26 &  36.56 &  70.70 &  36.82 &  32.89 &   0.60 &   0.60 &   0.00 \\
&    5 &  78.00 &  46.24 &  37.99 &  \textbf{73.10} &  47.32 &  30.62 &   0.70 &   0.62 &   0.04 \\
&    6 &  74.60 &  44.76 &  35.72 &  71.30 &  \textbf{64.10 }&   8.46 &   0.70 &   0.68 &   0.04 \\
&    7 &  76.30 &  46.04 &  37.15 &  67.30 &  46.34 &  21.95 &   0.70 &   0.62 &   0.04 \\
&    8 &  74.00 &  44.52 &  36.66 &  65.20 &  47.18 &  18.95 &   0.70 &   0.62 &   0.04 \\
&    9 &  \textbf{78.60} &  \textbf{57.46} &  28.51 &  62.70 &  54.46 &   9.82 &   0.70 &   0.64 &   0.05 \\
&   10 &  76.20 &  31.60 &  38.03 &  67.20 &  42.28 &  24.59 &   0.60 &   0.60 &   0.00 \\ \midrule
&    avg &  75.66 &  38.30 &  35.01 &  65.79 &  42.91 &  22.19 &   0.69 &   0.63 &   0.03 \\ \midrule
&   avg w/o Cos &  77.02 &  46.73 &  31.90 &   4.07 &   3.05 &   0.74 &   0.86 &   0.68 &   0.11 \\ 
&   avg w/o KL  &   2.41 &   2.01 &   0.28 &   3.36 &   2.60 &   0.57 &   0.79 &   0.69 &   0.07 \\
\bottomrule
\end{tabular}
}
    \caption{Matching accuracy with continuous and discrete reversal tokens in different settings. \textbf{\#rt} refer to the number of reversal tokens. We report the max and mean accuracy with standard deviation across 5 seeds. For comparison, we report matching accuracy between prompts with/without IKE-edits in the first top row after the header. }
    \label{tab:match_acc}
\end{table*}

\section{Analysis}
\label{sec:analysis}

\begin{table*}[!h]
    \centering
    \small
\begin{tabular}{lccccccccc}
\toprule
\tabhead
 & \multicolumn{3}{c}{\textbf{GPT2-XL}} & \multicolumn{3}{c}{\textbf{GPT-J}} & \multicolumn{3}{c}{\textbf{LLAMA-3.1-8B}} \\

\textbf{Setting} &  \textbf{Max} &  \textbf{Mean} &  \textbf{Std} &   \textbf{Max} &  \textbf{Mean} &  \textbf{Std} &   \textbf{Max} &  \textbf{Mean} &  \textbf{Std} \\
\midrule

   Continuous &  86.19 &  84.28 &   1.74 &  87.56 &  86.29 &   1.10 &  92.62 &  91.48 &   1.17 \\

    Discrete &  79.44 &  74.57 &   4.01 &  74.85 &  69.74 &   4.17 &  79.73 &  75.70 &   3.42   \\
\bottomrule
\end{tabular}
    \caption{Max and mean matching accuracy between normal (unedited) prompts \emph{with} and \emph{without} reversal tokens. Ideally, reversal tokens should not affect the outputs of normal prompts if there are no IKE-edits. Continuous reversal tokens have less effects on normal prompts (higher accuracy) compared to discrete reversal tokens.  }
    \label{tab:match_acc:normal}
\end{table*}

\begin{CJK*}{UTF8}{gbsn}
\begin{table}[h!]
    \centering
    \small
\begin{tabular}{cccccc}

\toprule
 \multicolumn{2}{c}{\textbf{GPT2-XL}} & \multicolumn{2}{c}{\textbf{GPT-J}} & \multicolumn{2}{c}{\textbf{LLAMA-3.1-8B}} \\
 Acc. & Tokens & Acc. & Tokens & Acc. & Tokens  \\ \midrule
  3.8 &          UCLA &   3.5 &             a &   0.7 &      ümü \\
  4.1 &       NETWORK &   3.5 &             a &   0.7 &        ? \\
  3.5 &    cellaneous &  61.1 & \bos &   0.7 &      \_ST \\
  3.7 &         Henry &   3.5 &             ( &   0.7 &        й \\
 75.7 & \bos &   4.0 &             d &   0.9 &  Revised \\

\bottomrule
\end{tabular}
    \caption{The found discrete reversal token. The \bos token helps reverse edits in GPT-models. }%
    \label{tab:disc_tokens}
\end{table}

\end{CJK*}

\begin{figure*}[]
\centering
\includegraphics[width=\textwidth, trim={3cm 0cm 12cm 0cm}, clip]{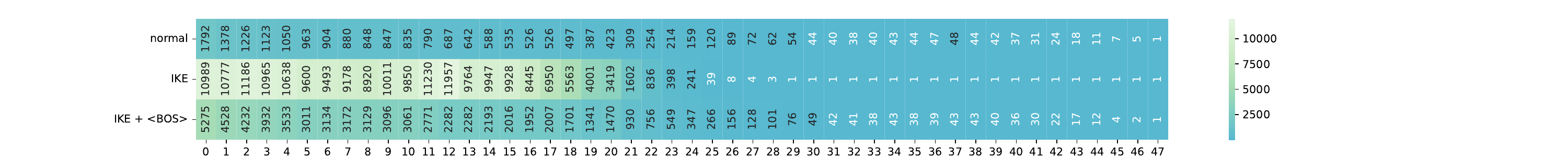}\\
  
\caption{Ranking of the final output over different layers in GPT2-XL. Rankings are averaged and rounded over 2000 instances. IKE causes the model to become certain about its prediction at earlier layers, while the model normally (no IKE prompt) keeps on refining its predictions until the final layer.}%
\label{fig:rankings}
\end{figure*}

We analyze the found reversal tokens under the discrete setting. Our goal is to investigate which \emph{natural} tokens help reverse edits. We show examples with 1 reversal token in Tab.~\ref{tab:disc_tokens}, and more examples in Fig.~\ref{fig:discrete_tokens} in the Appendix. The found tokens contain some non-English tokens. In GPT2-XL and GPT-J, we notice that the prompts that contain the beginning of sentence token (\bos) achieve high accuracy (more than 70\% and 60\% in GPT2-XL and GPT-J respectively). The \bos token is often used at the beginning of training instances and signals the onset of a new context. It is therefore intuitive for the \bos to be helpful in reversing edits. We experimented with the \bos token in LLAMA3.1, but  this did not lead to improved performance. This might be due to differences in the training procedure in LLAMA3.1 compared to the GPT-models.

\paragraph{Effect of reversal tokens on normal prompts.} Reversal tokens should restore the original outputs when IKE-edits are present in the prompt, but ideally should not affect the outputs if the prompts do not contain IKE-edits. To assess this, we compare the matching accuracy between normal prompts with and without reversal tokens ($r \oplus p(s,r)$ vs. $p(s,r)$). Results in Tab.~\ref{tab:match_acc:normal} show that continuous tokens change only 9-15\% of the outputs on average, while discrete tokens affect 25-30\%. This shows that continuous tokens have less side-effects. These unintended changes should be further minimized, but can be mitigated by first detecting an edit and applying reversal tokens only when edits are suspected. We encourage developing more robust methods in future work.

\paragraph{Fact retrieval with IKE.} Our analysis aims to understand how IKE affects fact retrieval in GPT2-XL, and how adding \bos as reversal token impacts this process. We analyze fact retrieval under three conditions: without IKE edits, with IKE edits, and with IKE edits plus the \bos token. We project output representations from each GPT2-XL layer into the vocabulary space~\cite{logitlens, alammar-2021-ecco}.\footnote{In Transformers, representations can be viewed as an information stream, with each layer reading from and writing back to this stream~\cite{elhage2021mathematical, geva-etal-2022-transformer}.} Projecting the output representation at each layer gives us a probability distribution over the vocabulary, which represents the model's output at this layer. Specifically, we show the rank of the final output at each layer, indicating the probability evolution of the final output throughout the model. 

Fig.~\ref{fig:rankings} shows that the ranks of the final outputs with no IKE-edits keep on changing until the final layer, i.e., the model is making use of all of its layers to refine the final prediction. Conversely, with IKE-edits the rank of the final output does not change after layer 29, i.e., the model is certain about its output much earlier compared to the normal case. One can also notice that the final output is less probable at earlier layers with IKE. We attribute this to the longer inputs that the model needs to process in the IKE setting. Adding \bos makes the rankings with IKE more similar to the normal rankings, where the predictions are refined until the last layer. At early layers, we notice the final prediction becomes more probable after adding \bos (ranks 5275 instead of 10989), which suggests that \bos helps the model filter out potential outputs that come from the IKE prompt.

\begin{figure}[]
\centering
\includegraphics[width=0.85\columnwidth, trim={.7cm 1.05cm 2cm 1.8cm}, clip]{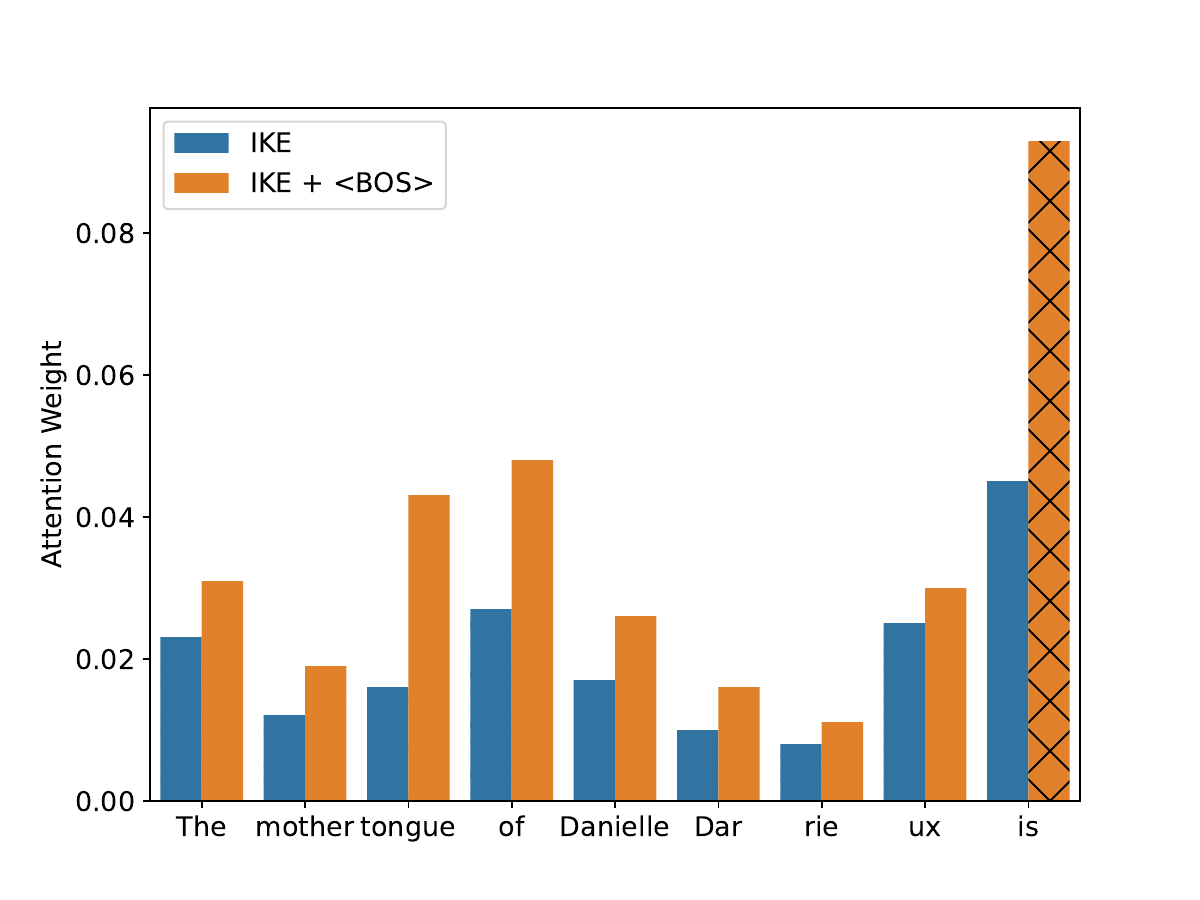}\\

\caption{Attention weights on an example from \cfd with an IKE-edit. Adding \bos helps the model to focus more on the query $p(s,r)$. IKE part of the input is omitted for space reasons.}%
\label{fig:att_example}
\end{figure}

\paragraph{Attention weights.} We further investigate how adding the \bos token affects the attention weights distribution in GPT2-XL~\cite{bahdanau2016neuralmachinetranslationjointly, sarti-etal-2023-inseq}. More specifically, we consider the averaged attention weights that are assigned to the query $p(s,r)$ in two settings: 1) IKE-edited; 2) IKE-edited with \bos over a sample of 100 examples. We hypothesize that adding \bos encourages the model to ignore the IKE-edit (prior context) in the prompt and focus more on $p(s,r)$, i.e., assign $p(s,r)$ higher attention weights. Indeed, we observe that when adding \bos the attention mass increases from 15\% to 38\% of the original attention mass observed in the unedited setting, despite the large increase in the input length from 8.13 to 877.3 tokens on average with IKE-edits. This suggests that \bos helps the model to ignore the IKE-edit and to focus more on the subsequent query. We show the attention weights for one example in Fig.~\ref{fig:att_example}. Adding \bos clearly leads to higher attention weights for the query.

\section{Conclusion}

In this work, we studied detecting and reversing IKE-edits in LLMs. Our work demonstrated that it is possible to detect IKE-edits with high accuracy, even in scenarios where the model's parameters are not changed and where only limited information about the model's outputs is available. We further showed that IKE-edits can be neutralized by tuning special reversal tokens.
By achieving over 80\% accuracy in recovering original outputs, continuous reversal tokens offer a promising method for mitigating the effects of malicious edits, and making LLMs intrinsically more secure. The effectiveness of continuous reversal tokens across different LLMs suggests a degree of generalizability that could be valuable for widespread implementation.
We believe our work contributes significantly to the ongoing efforts to make LLMs more secure, transparent, and trustworthy. %
Our future research will focus on improving the accuracy and robustness of both continuous and discrete reversal tokens.

\section{Limitations}

Following~\citet{zheng-etal-2023-edit}, we experimented with \cfd{} only, since IKE-edits require demonstrations that share the same relation as the edited facts, and such relations/demonstrations are not easily identifiable in other datasets. As a first venture for studying reversing edits, we assumed white-box access to LLMs to add and tune special reversal tokens. However, such access might be difficult to attain for users in practice.

\section*{Acknowledgment}
This research was partially funded by the German Federal Ministry of Education and Research (BMBF) as German Academic Exchange Service (DAAD) project DEED under Grant No. 30001797.

\bibliography{custom}
\bibliographystyle{acl_natbib}

\appendix

\section{Additional Results}
\label{sec:appendix}
\begin{figure*}[]
\centering
\includegraphics[width=\textwidth, trim={3cm 23cm 3cm 2cm}, clip]{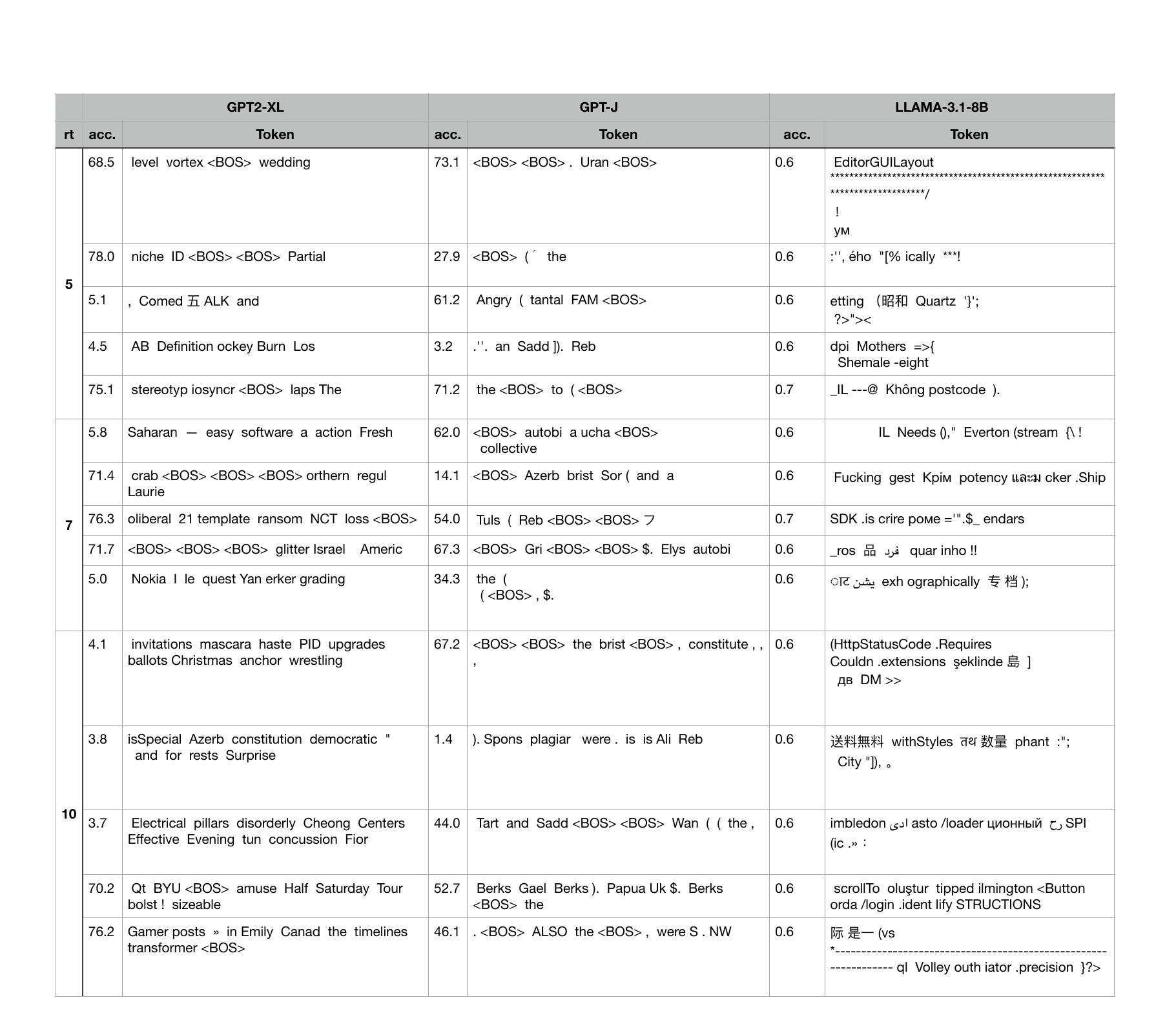}\\
\caption{The found discrete tokens with  5, 7 and 10 reversal tokens and 5 seeds. }%
\label{fig:discrete_tokens}
\end{figure*}

We show the learned reversal tokens in Fig.~\ref{fig:discrete_tokens}. Tab.~\ref{tab:match_acc:app} is a more comprehensive version of Tab.~\ref{tab:match_acc} (with varying numbers of reversal tokens for the ablation settings).

\paragraph{Generalization to other datasets.} We use MQuAKE~\cite{zhong-etal-2023-mquake}, which contains multi-hop edits to study the generalization of the proposed continuous reversal tokens. Since this dataset does not contain demonstration prompts for IKE, we follow \citet{huang2024knowledgeeditingreallycorrect} in using prompts that contain the new facts for editing (see example in Fig.~\ref{fig:mquake_example}). The proposed continuous reversal tokens generalize well to MQuAKE, showing improved performance with GPT models but decreased performance with LLAMA. We attribute this to MQuAKE's multi-hop edits, which challenge larger models to resist producing the edited answer. Despite the positive results, we believe the absence of demonstration prompts does not allow us to perfectly simulate the IKE setting, which we focus on in this paper.

\begin{table*}[]
    \centering
    \small
\begin{tabular}{lcccccccccc}
\toprule
\tabhead
& & \multicolumn{3}{c}{\textbf{GPT2-XL}} & \multicolumn{3}{c}{\textbf{GPT-J}} & \multicolumn{3}{c}{\textbf{LLAMA-3.1-8B}} \\

& \textbf{\#rt} &  \textbf{Max} &  \textbf{Mean} &  \textbf{Std} &   \textbf{Max} &  \textbf{Mean} &  \textbf{Std} &   \textbf{Max} &  \textbf{Mean} &  \textbf{Std} \\
\midrule
&   IKE vs. no-IKE &   1.90 &   1.90 &   0.00 &   2.20 &   2.20 &   0.00 &   0.60 &   0.60 &   0.00 \\ \midrule
   \multirow{10}{*}{\rotatebox[origin=c]{90}{\textbf{Cont.} Tuning}}
&    1.0 &  77.60 &  57.82 &  22.36 &  80.80 &  79.48 &   0.95 &  83.90 &  82.42 &   0.88 \\
&    2.0 &  79.10 &  61.96 &  21.33 &  80.80 &  79.42 &   0.99 &  85.20 &  82.82 &   1.48 \\
&    3.0 &  75.00 &  44.98 &  21.38 &  80.90 &  79.92 &   0.96 &  84.80 &  83.24 &   1.32 \\
&    4.0 &  78.80 &  62.80 &  22.28 &  81.80 &  80.56 &   0.78 &  84.90 &  83.48 &   1.01 \\
&    5.0 &  79.20 &  71.64 &  14.77 &  81.30 &  80.38 &   0.67 &  84.80 &  83.88 &   0.90 \\
&    6.0 &  79.60 &  69.08 &  14.09 &  81.60 &  80.90 &   0.61 &  84.90 &  83.34 &   0.93 \\
&    7.0 &  82.10 &  75.52 &   7.85 &  82.20 &  81.44 &   0.61 &  85.70 &  84.12 &   1.80 \\
&    8.0 &  82.30 &  \textbf{78.14} &   4.96 &  83.10 &  81.74 &   1.44 &  85.10 &  84.48 &   0.77 \\
&    9.0 &  \textbf{82.90} &  73.70 &   6.16 &  83.00 &  81.94 &   0.91 &  \textbf{86.00} &  \textbf{84.64} &   1.05 \\
&   10.0 &  80.10 &  71.56 &  12.73 &  \textbf{84.30} &  \textbf{82.52} &   1.10 &  85.40 &  84.40 &   1.03 \\ \midrule
&    avg &  79.67 &  66.72 &  14.79 &  81.98 &  80.83 &   0.90 &  85.07 &  83.68 &   1.12 \\ \midrule
   \multirow{10}{*}{\rotatebox[origin=c]{90}{\textbf{Disc.} Tuning}}

&    1.0 &  75.70 &  18.16 &  32.17 &  61.10 &  15.12 &  25.70 &   \textbf{0.90} &   \textbf{0.74} &   0.09 \\
&    2.0 &  73.10 &  43.30 &  35.75 &  60.50 &  33.30 &  26.89 &   0.70 &   0.62 &   0.04 \\
&    3.0 &  75.10 &  18.68 &  31.55 &  58.80 &  42.20 &  22.03 &   0.60 &   0.60 &   0.00 \\
&    4.0 &  75.00 &  32.26 &  36.56 &  70.70 &  36.82 &  32.89 &   0.60 &   0.60 &   0.00 \\
&    5.0 &  78.00 &  46.24 &  37.99 &  \textbf{73.10} &  47.32 &  30.62 &   0.70 &   0.62 &   0.04 \\
&    6.0 &  74.60 &  44.76 &  35.72 &  71.30 &  \textbf{64.10} &   8.46 &   0.70 &   0.68 &   0.04 \\
&    7.0 &  76.30 &  46.04 &  37.15 &  67.30 &  46.34 &  21.95 &   0.70 &   0.62 &   0.04 \\
&    8.0 &  74.00 &  44.52 &  36.66 &  65.20 &  47.18 &  18.95 &   0.70 &   0.62 &   0.04 \\
&    9.0 &  \textbf{78.60} &  \textbf{57.46} &  28.51 &  62.70 &  54.46 &   9.82 &   0.70 &   0.64 &   0.05 \\
&   10.0 &  76.20 &  31.60 &  38.03 &  67.20 &  42.28 &  24.59 &   0.60 &   0.60 &   0.00 \\ \midrule
&    avg &  75.66 &  38.30 &  35.01 &  65.79 &  42.91 &  22.19 &   0.69 &   0.63 &   0.03 \\ \midrule
   \multirow{10}{*}{\rotatebox[origin=c]{90}{\textbf{Disc.} $-$ Cos loss}}
&    1.0 &  75.70 &  32.48 &  39.47 &   5.90 &   4.70 &   1.07 &   0.70 &   0.64 &   0.05 \\
&    2.0 &  74.90 &  32.46 &  37.62 &   \textbf{8.00} &   4.88 &   1.94 &   \textbf{1.90} &   \textbf{0.92} &   0.55 \\
&    3.0 &  78.60 &  18.76 &  33.49 &   5.80 &   \textbf{5.02} &   0.48 &   1.00 &   0.72 &   0.16 \\
&    4.0 &  74.60 &  45.84 &  39.06 &   3.80 &   3.00 &   0.61 &   0.70 &   0.66 &   0.05 \\
&    5.0 &  78.50 &  \textbf{71.56} &   8.69 &   2.50 &   2.00 &   0.52 &   0.70 &   0.66 &   0.05 \\
&    6.0 &  78.60 &  32.32 &  39.13 &   2.30 &   1.96 &   0.36 &   0.70 &   0.66 &   0.05 \\
&    7.0 &  78.70 &  60.30 &  29.48 &   3.10 &   2.36 &   0.57 &   0.80 &   0.64 &   0.09 \\
&    8.0 &  75.60 &  56.28 &  30.16 &   2.90 &   2.00 &   0.60 &   0.60 &   0.60 &   0.00 \\
&    9.0 &  76.20 &  58.10 &  31.12 &   2.90 &   2.20 &   0.56 &   0.80 &   0.66 &   0.09 \\
&   10.0 &  \textbf{78.80} &  59.20 &  30.78 &   3.50 &   2.38 &   0.70 &   0.70 &   0.62 &   0.04 \\ \midrule
&   avg  &  77.02 &  46.73 &  31.90 &   4.07 &   3.05 &   0.74 &   0.86 &   0.68 &   0.11 \\ \midrule

   \multirow{10}{*}{\rotatebox[origin=c]{90}{\textbf{Disc.} $-$ KL loss}}

&  1.0 &   2.00 &   2.00 &   0.00 &   3.50 &   \textbf{3.50} &   0.00 &   0.60 &   0.60 &   0.00 \\
&  2.0 &   2.00 &   1.58 &   0.24 &   3.30 &   2.38 &   0.52 &   0.70 &   0.62 &   0.04 \\
&  3.0 &   2.20 &   1.96 &   0.23 &   3.00 &   2.44 &   0.59 &   \textbf{1.20} &   \textbf{0.78} &   0.25 \\
&  4.0 &   2.50 &   1.94 &   0.48 &   2.90 &   2.36 &   0.52 &   0.70 &   0.70 &   0.00 \\
&  5.0 &   2.50 &   2.06 &   0.36 &   2.90 &   2.38 &   0.46 &   0.80 &   0.72 &   0.04 \\
&  6.0 &   2.80 &   \textbf{2.28} &   0.34 &   3.00 &   2.64 &   0.35 &   0.70 &   0.64 &   0.05 \\
&  7.0 &   2.40 &   1.94 &   0.27 &   \textbf{5.40} &   2.94 &   1.39 &   0.80 &   0.70 &   0.07 \\
&  8.0 &   2.50 &   2.06 &   0.30 &   2.90 &   2.36 &   0.48 &   0.70 &   0.66 &   0.05 \\
&  9.0 &   2.20 &   1.92 &   0.18 &   3.30 &   2.46 &   0.63 &   0.80 &   0.68 &   0.08 \\
& 10.0 &   \textbf{3.00} &   2.40 &   0.40 &   3.40 &   2.50 &   0.78 &   0.90 &   0.76 &   0.11 \\ \midrule
&  avg &   2.41 &   2.01 &   0.28 &   3.36 &   2.60 &   0.57 &   0.79 &   0.69 &   0.07 \\
\bottomrule
\end{tabular}
    \caption{Matching accuracy with continuous and discrete reversal tokens in different settings. \textbf{\#rt} refer to the number of reversal tokens. We report the max and mean accuracy with standard deviation across 5 seeds. For comparison, we report matching accuracy between prompts with/without IKE-edits in the first top line. }
    \label{tab:match_acc:app}
\end{table*}

\begin{table*}[]
    \centering
    \small
\begin{tabular}{lcccccccccc}
\toprule
\tabhead
& & \multicolumn{3}{c}{\textbf{GPT2-XL}} & \multicolumn{3}{c}{\textbf{GPT-J}} & \multicolumn{3}{c}{\textbf{LLAMA-3.1-8B}} \\

& \textbf{\#rt} &  \textbf{Max} &  \textbf{Mean} &  \textbf{Std} &   \textbf{Max} &  \textbf{Mean} &  \textbf{Std} &   \textbf{Max} &  \textbf{Mean} &  \textbf{Std} \\
\midrule
&   IKE vs. no-IKE &   1.00 &   1.00 &   0.00 &    0.5 &   0.50 &   0.00 &   0.20 &   0.20 &   0.00 \\  \midrule
   \multirow{10}{*}{\rotatebox[origin=c]{90}{\textbf{Cont.} Tuning}}
&   1.0 &  99.10 &  78.66 &  42.45 &  100.0 & 100.00 &   0.00 &  61.80 &  57.96 &   2.71 \\
&   2.0 &  99.10 &  98.82 &   0.28 &  100.0 & 100.00 &   0.00 &  65.60 &  61.90 &   3.14 \\
&   3.0 & 100.00 &  97.70 &   2.60 &  100.0 &  99.98 &   0.04 &  63.80 &  60.88 &   3.25 \\
&   4.0 & 100.00 &  98.10 &   3.29 &  100.0 &  99.98 &   0.04 &  67.60 &  62.96 &   2.60 \\
&   5.0 & 100.00 &  99.10 &   1.00 &  100.0 &  99.98 &   0.04 &  66.20 &  63.10 &   2.09 \\
&   6.0 &  99.80 &  98.58 &   1.63 &  100.0 & 100.00 &   0.00 &  67.90 &  64.78 &   2.36 \\
&   7.0 & 100.00 &  99.80 &   0.14 &  100.0 & 100.00 &   0.00 &  67.40 &  65.06 &   1.92 \\
&   8.0 &  99.90 &  95.36 &   9.87 &  100.0 & 100.00 &   0.00 &  68.50 &  65.68 &   1.95 \\
&   9.0 &  99.70 &  99.52 &   0.20 &  100.0 &  99.96 &   0.09 &  66.60 &  65.62 &   1.14 \\
&  10.0 &  99.90 &  94.12 &  12.48 &  100.0 &  99.98 &   0.04 &  68.70 &  67.22 &   1.55 \\ \midrule
&   avg &  99.75 &  95.98 &   7.39 &  100.0 &  99.99 &   0.02 &  66.41 &  63.52 &   2.27 \\

\bottomrule
\end{tabular}
    \caption{Matching accuracy with continuous reversal tokens using MQuAKE. \textbf{\#rt} refer to the number of reversal tokens. We report the max and mean accuracy with standard deviation across 5 seeds. For comparison, we report matching accuracy between prompts with/without IKE-edits in the first top line. }
    \label{tab:match_acc_mquake:app}
\end{table*}

\begin{figure*}[]
\centering
  \includegraphics[scale=0.3, trim={0cm 0cm 0cm 1.30cm}, clip]{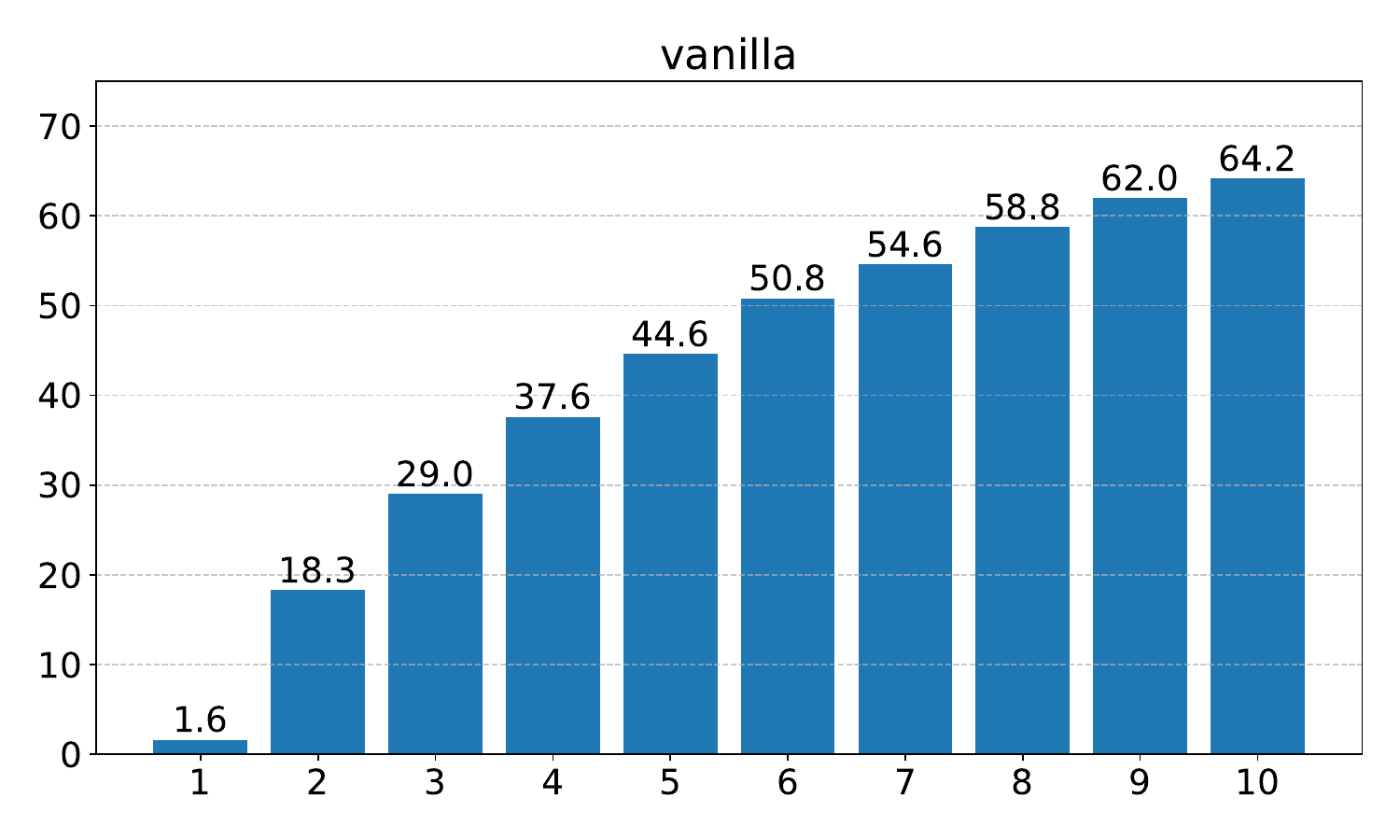}\quad
  \includegraphics[scale=0.3, trim={0cm 0cm 0cm 1.30cm}, clip]{figs/gpt-j-6B_vanilla.pdf}
\caption{Percentage of top1 outputs \emph{before} editing that can be found in the top10 outputs of the model \emph{after} editing (cumulative histogram). Left: GPT2-XL, right: GPT-J. } %
\label{fig:output_shift:app}
\end{figure*}

\begin{figure*}
\setlength\fboxsep{7pt}
\fbox{%
\parbox{0.95\linewidth}{%
\begin{courier}

New Fact: The mother tongue of Jonathan Littell is Greek
Prompt: Jonathan Littell, speaker of Greek

New Fact: The mother tongue of Michel Braudeau is Russian
Prompt: Michel Braudeau spoke the language Russian

New Fact: The mother tongue of Louis Florencie is Russian
Prompt: Louis Florencie spoke the language Russian

New Fact: The mother tongue of Rainer Maria Rilke is French
Prompt: Moritz Cantor spoke the language German

New Fact: The mother tongue of Robert Lecourt is English
Prompt: Robert Lecourt, a native English

New Fact: The mother tongue of Jan Wils is Italian
Prompt: Johan Daisne is a native speaker of Dutch

New Fact: The mother tongue of Elsa Zylberstein is German
Prompt: Elsa Zylberstein spoke the language German

New Fact: The mother tongue of Daniel-Rops is Polish
Prompt: The native language of Montesquieu is French

New Fact: The mother tongue of Jan Commelin is French
Prompt: Henk van Woerden spoke the language Dutch

New Fact: The mother tongue of Alain Marleix is Russian
Prompt: The native language of Montesquieu is French

New Fact: The mother tongue of Jean-Baptiste Solignac is Russian
Prompt: François Bayrou, speaker of French

New Fact: The mother tongue of Michel d'Ornano is Swedish
Prompt: The native language of Michel d'Ornano is Swedish

New Fact: The mother tongue of Ewa Farna is French
Prompt: Bernard Czapliński, a native Polish

New Fact: The mother tongue of Patrick Chamoiseau is Dutch
Prompt: Melchior de Vogüé, speaker of French

New Fact: The mother tongue of Dominique Cabrera is Dutch
Prompt: The mother tongue of Dominique Cabrera is Dutch

New Fact: The mother tongue of Henri Diamant-Berger is English
Prompt: Jean Auguste Dominique Ingres spoke the language French

New Fact: The mother tongue of Dominique Zardi is Dutch
Prompt: The mother tongue of Dominique Zardi is Dutch

New Fact: The mother tongue of Michel Camdessus is Russian
Prompt: Robert Schuman, a native French

New Fact: The mother tongue of Catherine Picard is Dutch
Prompt: Catherine Picard is a native speaker of Dutch

New Fact: The mother tongue of Martin Lamotte is Latin
Prompt: Jean Auguste Dominique Ingres, speaker of French

New Fact: The mother tongue of Philippe de Mornay is Russian
Prompt: Georges Duhamel, a native French

New Fact: The mother tongue of Marie NDiaye is Russian
Prompt: Marie NDiaye is a native speaker of Russian

New Fact: The mother tongue of Jean-Antoine Chaptal is English
Prompt: Léon Blum is a native speaker of French

New Fact: The mother tongue of Catherine Deneuve is Dutch
Prompt: The mother tongue of Catherine Deneuve is Dutch

New Fact: The mother tongue of Raymond Triboulet is Dutch
Prompt: Jean Gabin, a native French

New Fact: Daniel Darc is a native speaker of Dutch
Prompt: Léon Blum is a native speaker of French

New Fact: The mother tongue of Louis Carrogis Carmontelle is Polish
Prompt: Louis Carrogis Carmontelle spoke the language Polish

New Fact: The mother tongue of Daniel Pennacchioni is Russian
Prompt: The native language of Daniel Pennacchioni is Russian

New Fact: The mother tongue of Camille Flammarion is Dutch
Prompt: Camille Flammarion, speaker of Dutch

New Fact: The mother tongue of Bernard Cerquiglini is English
Prompt: Henri Barbusse, speaker of French

New Fact: The mother tongue of Marc-Philippe Daubresse is Russian
Prompt: The mother tongue of Marc-Philippe Daubresse is Russian

New Fact: The mother tongue of Colette Darfeuil is Russian
Prompt: Colette Darfeuil spoke the language Russian

New Fact: The mother tongue of Danielle Darrieux is English
Prompt: The mother tongue of Danielle Darrieux is English

\end{courier}
}}
\caption{An example of $p_{edit}(s,r,o')$ that changes the mother tongue of Danielle Darrieux from French to English.}
\label{fig:ike_example}
\end{figure*}

\begin{figure*}
\setlength\fboxsep{7pt}
\fbox{%
\parbox{0.95\linewidth}{%
\begin{courier}
Who is the original broadcaster of Grey's Anatomy? British Broadcasting Corporation
Who is the director of British Broadcasting Corporation? Narendra Modi
What is the country of citizenship of Narendra Modi? Australia
What is the capital of Australia? Oderzo
Which city is the capital of the country where the director/manager of Grey's Anatomy was born? Oderzo %
\end{courier}
}}
\caption{An example of $p_{edit}(s,r,o')$ from MQuAKE. Due to the lack of IKE demonstrations, we use prompts that contain the answers.}
\label{fig:mquake_example}
\end{figure*}

\section{Further Implementation Details}
\label{sec:app_imp}
\paragraph{Dimensions for resetting context.} For step 1 in discrete prompt tuning, we experimented with 5  random seeds, and continued the experiments with the seed whose accuracy corresponded to the median accuracy value.

\section{Computational Resources}
\label{sec:app_compute}
All of our experiments were conducted with an NVIDIA A100 GPU with 80GB of memory. Our experiments took roughly 10 GPU days.

\end{document}